\documentclass[final]{cvpr}
\usepackage{tabulary,xspace}
\usepackage{tabularx}
\usepackage{times}
\usepackage{epsfig}
\usepackage{graphicx}
\usepackage{fixmath,mathtools,nicefrac,mmstyle}
\usepackage[table,xcdraw]{xcolor}
\usepackage{amsmath}
\usepackage{amssymb}
\usepackage{caption}
\usepackage{multirow}
\usepackage{booktabs}
\usepackage[section]{placeins}
\usepackage{float}
\usepackage{wrapfig}

\usepackage[pagebackref,breaklinks,colorlinks]{hyperref}

\definecolor{ourscolor}{gray}{.9}

\newcommand{\authorskip}{\hspace{2.5mm}}

\newcommand\ntfootnote[1]{%
  \begin{NoHyper}
  \renewcommand\thefootnote{}\footnotetext{#1}%
  \addtocounter{footnote}{0}%
  \end{NoHyper}
}

\def\cvprPaperID{****} 
\def\confName{CVPR}
\def\confYear{2021}
\graphicspath{{./figs/}{./figs/result/}}

\begin{document}

\title{RTMW: Real-Time Multi-Person 2D and 3D Whole-body Pose Estimation}

\author{Tao Jiang$^{\ast}$ \authorskip Xinchen Xie$^{\ast}$ \authorskip Yining Li \\[2mm]
Shanghai AI Laboratory \\
{\tt\small $\left \{\text{jiangtao, xiexinchen, liyining}\right\}$@pjlab.org.cn} \\
}

\maketitle
\ntfootnote{$^{\ast}$ Equal Contribution.}


\begin{abstract}

   Whole-body pose estimation is a challenging task that requires simultaneous prediction of keypoints for the body, hands, face, and feet.
   Whole-body pose estimation aims to predict fine-grained pose information for the human body, including the face, torso, hands, and feet, which plays an important role in the study of human-centric perception and generation and in various applications.
   In this work, we present RTMW (Real-Time Multi-person Whole-body pose estimation models), a series of high-performance models for 2D/3D whole-body pose estimation.
   We incorporate RTMPose model architecture with FPN and HEM (Hierarchical Encoding Module) to better capture pose information from different body parts with various scales. The model is trained with a rich collection of open-source human keypoint datasets with manually aligned annotations and further enhanced via a two-stage distillation strategy.
   RTMW demonstrates strong performance on multiple whole-body pose estimation benchmarks while maintaining high inference efficiency and deployment friendliness. We release three sizes: m/l/x, with RTMW-l achieving a 70.2 mAP on the COCO-Wholebody benchmark, making it the first open-source model to exceed 70 mAP on this benchmark. Meanwhile, we explored the performance of RTMW in the task of 3D whole-body pose estimation, conducting image-based monocular 3D whole-body pose estimation in a coordinate classification manner.
   We hope this work can benefit both academic research and industrial applications. The code and models have been made publicly available at: \url{https://github.com/open-mmlab/mmpose/tree/main/projects/rtmpose}

\end{abstract}

\section{Introduction}\label{sec:Introduction}

Whole-body pose estimation is an essential component in advancing the capabilities of human-centric artificial intelligence systems. It can be utilized in human-computer interaction, virtual avatar animation, and the film industry. In the context of AIGC (AI Generated Content) applications, the outcomes of whole-body pose estimation are also employed to control character generation. As the emergence of downstream tasks and industrial applications empowered by whole-body pose estimation, designing a model that is highly accurate, low-latency, and easy to deploy is extremely valuable.

In early research on human pose estimation, due to the complexity of the task and limitations in computational power and data, researchers divided the human body into separate parts for independent pose estimation studies. With the relentless efforts of predecessors, remarkable results have been achieved in these singular, part-specific 2D pose estimation tasks.

Previous work, such as OpenPose~\cite{openpose}, could obtain whole-body pose estimation results by combining the outcomes of these separate parts. However, this straightforward combination approach faced high computational costs and significant performance limitations. While lightweight tools like MediaPipe~\cite{mediapipe2019} offer high real-time performance and ease of deployment, their accuracy is not entirely satisfactory.

Our MMPose~\cite{mmpose2020} team released the RTMPose~\cite{https://doi.org/10.48550/arxiv.2303.07399} model last year, which achieved an excellent balance between accuracy and real-time performance. Subsequently, on this foundation, DWPose~\cite{yang2023effective} further enhanced the performance of RTMPose in whole-body pose estimation tasks by incorporating a two-stage distillation technique and integrating a new dataset, UBody~\cite{lin2023one}.

The structural design of RTMPose~\cite{https://doi.org/10.48550/arxiv.2303.07399} initially only considered the body posture. However, in whole-body pose estimation tasks, feature resolution is crucial for accuracy in facial, hand, and foot pose estimation. Therefore, we introduced two techniques, PAFPN (Part-Aggregation Feature Pyramid Network) and HEM (High-Efficiency Multi-Scale Feature Fusion), to enhance feature resolution. Experimental results have confirmed that these two modules significantly improve the localization accuracy of fine-grained body parts.

At the same time, the scarcity of open-source whole-body pose estimation datasets greatly limits the performance of open-source models. To fully use datasets focusing on different body parts, we manually aligned the key point definitions of 14 open-source datasets (3 for whole-body keypoints, 6 for body keypoints, 4 for facial keypoints, and 1 for hand keypoints), which are jointly used to train RTMW.

In the realm of 3D whole-body pose estimation, the academic community has predominantly embraced two main methodologies: Lifting~\cite{martinez_2017_3dbaseline, pavllo:videopose3d:2019, motionbert2022} and regression~\cite{pavlakos17volumetric, sun2017integral} approaches. There has been a notable absence of scholarly inquiry into methods grounded in the SimCC~\cite{SimCC} technique. Our research endeavors have ventured into uncharted territory by applying the RTMW architecture to the 3D whole-body pose estimation task. Our experimental findings indicate that the SimCC~\cite{SimCC} method holds its own and delivers commendable performance in this domain.

\section{Related Work}


\paragraph{Top-down Approaches.} 
Top-down algorithms use off-the-shelf detectors to provide bounding boxes and crop the human to a uniform scale for pose estimation. Algorithms ~\cite{xiao2018simple, cai2020learning, SunXLW19, liu2021polarized, xu2022vitpose} of the top-down paradigm have dominated public benchmarks. The two-stage inference paradigm allows both the human detector and the pose estimator to use relatively small input resolutions, allowing them to outperform bottom-up algorithms in speed and accuracy in non-extreme scenarios (i.e. when the number of people in the image is no more than 6). Additionally, most previous work has focused on achieving state-of-the-art performance on public datasets. In contrast, our work aims to design models with better speed-accuracy trade-offs to meet the needs of industrial applications.

\paragraph{Coordinate Classification.} Previous pose estimation approaches usually regard keypoint localization as either coordinate regression (e.g. ~\cite{Toshev_2014_CVPR, li2021human, mao2022poseur}) or heatmap regression (e.g. ~\cite{xiao2018simple, huang2020devil, Zhang_2020_CVPR, xu2022vitpose}). SimCC~\cite{SimCC} introduces a new scheme that formulates keypoint prediction as classification from sub-pixel bins for horizontal and vertical coordinates respectively, which brings about several advantages. First, SimCC is freed from the dependence on high-resolution heatmaps, thus allowing for a very compact architecture that requires neither high-resolution intermediate representations ~\cite{SunXLW19} nor costly upscaling layers ~\cite{xiao2018simple}. Second, SimCC flattens the final feature map for classification instead of involving global pooling ~\cite{Toshev_2014_CVPR} and, therefore, avoids the loss of spatial information. Third, the quantization error can be effectively alleviated by coordinate classification at the sub-pixel scale without needing extra refinement post-processing ~\cite{Zhang_2020_CVPR}. These qualities make SimCC attractive for building lightweight pose estimation models. RTMO~\cite{lu2023rtmo} introduced the coordinate classification method into the one-stage pose estimation, which has achieved significant performance improvement and also confirmed the great potential of the SimCC method in pose estimation tasks. In this work, we further exploit the coordinate classification scheme with optimizations on model architecture and training strategy.

\paragraph{3D pose estimation}
3D pose estimation is a vibrant research domain with extensive industry applications. The landscape of contemporary methodologies is dominated by two primary approaches: lifting methods~\cite{martinez_2017_3dbaseline, pavllo:videopose3d:2019, motionbert2022} that leverage 2D keypoints and regression methods grounded in image analysis. The lifting methods, which input 2D coordinates into a neural network to directly predict their spatial coordinates, are distinguished by their swift computational pace. However, this efficiency comes at the cost of scene information, as these algorithms are devoid of image input, leading to a reliance on the annotation of training data for the range of their predictive outcomes.

Conversely, image-based regression methods~\cite{pavlakos17volumetric, sun2017integral}, while incorporating rich visual data, face the challenges of sluggish inference speeds and increased task complexity. These factors contribute to difficulty in training the models and achieving high accuracy. Our proposed method, RTMW3D, diverges from these conventional approaches by employing a classification strategy based on the Simcc technique~\cite{SimCC} to refine the final spatial coordinates through post-processing. Our experimental findings underscore the effectiveness of this innovative approach.


\section{Model Architecture and Training}

Although the RTMPose we proposed before did not have a special design for the whole-body keypoint estimation task, after experiments, we found that its performance is comparable to the current state-of-the-art method, ZoomNas~\cite{xu2022zoomnas}. However, during the experimental process, we found that RTMPose has a certain performance bottleneck in the whole-body pose estimation task. As the scale of parameters increases, the model performance does not improve with the increase in the number of parameters. On the other hand, with the research of other research teams, such as the DWPose team, although they added new datasets during the training of RTMPose and adopted more efficient two-stage distillation training technology, which effectively improved the accuracy of RTMPose, they still cannot avoid the problem of the inherent performance bottleneck of RTMPose as the number of parameters increases.

\subsection{RTMW}

\subsubsection{Task Limitation}

We first analyze some of the unsolved problems in the whole-body pose estimation task. RTMW was designed to address these challenges. The first issue is the resolution limit of local areas. In an image, parts of the human body, such as the face, hands, and feet, occupy a very small proportion of the human body and the image. For a model, the input resolution of these areas will directly affect the accuracy of the model's prediction of the keypoints in these parts. The second problem is that the difficulty of learning keypoints from different human body parts varies. For example, the keypoints on the face can be regarded as some points attached to the face, a rigid body, and since the deformation of the face is very small, the model will find it easier to learn to predict these facial keypoints. On the contrary, hand keypoints have a higher degree of freedom due to the rotation and movement of the fingers and wrists, making them much more difficult to predict.
The third problem lies in the loss function. Commonly used loss functions such as the KL divergence and the regression error are calculated point-by-point. Thus, they tend to be dominated by body parts with more keypoints and larger spatial proportions, like the torso and face, paying insufficient attention to small but complex parts like hands and feet. This typically results in imbalanced convergence, meaning the model archives lower average error but exhibits poor accuracy on complex body parts. The last point is that there are very few open-source whole-body pose estimation datasets, significantly limiting our research on the model's capabilities. In response to the above limitations, we have designed a set of targeted optimization solutions for RTMPose and proposed our RTMW model.

\subsubsection{Model Architecture}

\begin{figure*}[t]
    \centering
    \includegraphics[width=0.8\textwidth]{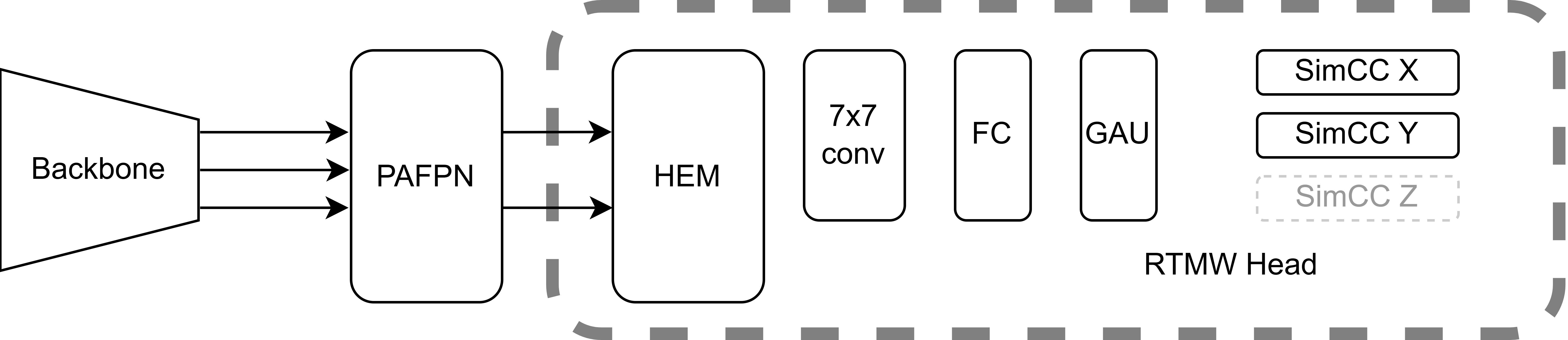}
    \caption{The RTMW arch.} 
    \label{fig:rtmw_model}
\end{figure*}

The RTMW model structure is shown in Figure~\ref{fig:rtmw_model}, designed based on RTMPose, incorporating PAFPN~\cite{liu2018path} and HEM modules to enhance feature resolution. Among these, FPN (Feature Pyramid Network) is an effective technique for improving feature resolution and is widely used in the model structures of dense prediction tasks. In addition to this, inspired by the hierarchical encoding concept proposed in the VQVAE-2~\cite{razavi2019generating} paper, we introduced a HEM (Hierarchical Encoding Module) module. Experimental results show that the simultaneous introduction of these two modules can significantly enhance the performance of RTMPose in whole-body pose estimation tasks.

\paragraph{PAFPN}
In the previous analysis, since we emphasized the impact of the input resolution of various body parts on the model's prediction accuracy, we naturally introduced an FPN module. This technique is very common in other vision tasks, such as object detection tasks. Details of this module can be referred to in the original paper~\cite{liu2018path} and another improved version used in RTMDet~\cite{lyu2022rtmdet}.

\paragraph{HEM (Hierarchical Encoding Module)}
The inspiration for this module comes from the work VQVAEv2, which introduced a hierarchical concept on top of the original Encoder-Decoder architecture. In the original VQVAE work, the generated images lacked clear and rich details. Therefore, in the second version, they added hierarchical encoding to features of different resolutions and decoded these hierarchical features separately during the decoding process, which enriched the details in the generated images. We were inspired by this idea in the design of RTMW, creating the HEM (Hierarchical Encoding Module), which performs SimCC encoding on the features output by PAFPN in a hierarchical manner, then merges them after the hierarchical encoding and finally decodes them in the decoder. Experiments have shown that this design can improve the prediction accuracy of low-resolution body parts in human pose estimation.

\subsubsection{Training Techniques}
Due to the lack of open-source whole-body pose estimation datasets, we manually aligned 14 open-source datasets that include whole-body, torso, hand, and facial keypoints for pose estimation. We used these 14 datasets for joint training. At the same time, to maximize the model's performance, we adopted the two-stage distillation technique used by DWPose during the model training process to further enhance the model's performance.

\subsection{RTMW3D}

We further explore extending RTMW architecture to 3D whole-body pose estimation, which is challenging due to its ill-posed nature. Specifically, while it is straightforward to ascertain the x and y coordinates of a keypoint from a single RGB image, the ambiguity of the z-axis introduces complexity in the model's learning process. Furthermore, the scarcity of open-source 3D human pose datasets, especially those with keypoint annotations, and the limited scene diversity of available datasets severely hamper the model's performance and generalization ability. To address these challenges, RTMW3D has been tailored to excel in monocular 3D pose estimation by refining the 3D dataset, redefining the z-axis, and mitigating the learning difficulty of predicting z-axis coordinates. Structurally, RTMW3D closely mirrors RTMW, with the notable addition of a z-axis prediction branch to the decoder head.

\begin{figure}[t]
    \centering
    \includegraphics[width=0.3\textwidth]{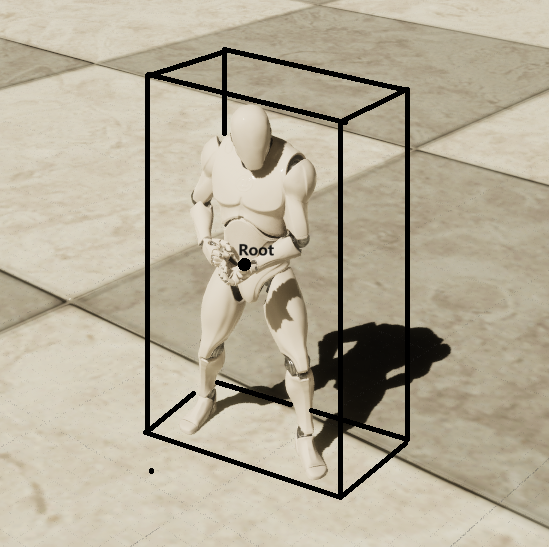}
    \caption{The task definition of 3d pose estimation.} 
    \label{fig:3d_example}
\end{figure}

\paragraph{Task definition}
RTMW3D adheres to the principles of 2D pose estimation, directly forecasting the x and y coordinates of keypoints on the SimCC coordinate space. Following the design of the SimCC~\cite{SimCC} method, as shown in figure~\ref{fig:3d_example}, for the z-axis, we eschew the direct use of annotated z-axis coordinates from the dataset. Instead, we establish a root point for the human skeleton and calculate the z-axis offsets of keypoints relative to this root point. This innovative approach standardizes the z-axis across various datasets and simplifies the model's learning challenge.

\paragraph{Data process}
Given the lack of 3D human pose datasets centered on keypoints, we implemented a synergistic training strategy that integrates both 2D and 3D datasets. This approach enriches the data with diverse scenarios, bolstering the model's performance on 2D pose estimation tasks. To address the issue of the lack of z-axis annotations in 2D keypoint datasets, we have incorporated a z-axis mask into the 2D keypoint data. This modification facilitates a unified training protocol for both data types, enabling the model to accurately predict the x and y coordinates of keypoints and enhance its capacity for 2D pose estimation.


\section{Experiments}\label{sec:Experiments}

\subsection{Datasets}

During the training phase, we continued to use all the training techniques employed in RTMPose and adopted the two-stage distillation technology proposed by DWPose. Due to the limited availability of open-source whole-body pose estimation datasets, we integrated 14 datasets from different parts, manually aligned the keypoint definitions, and uniformly mapped them to the 133-point definition of COCO-Wholebody. In the monocular 3D whole-body pose estimation task, due to the lack of open-source 3D datasets, we combined 14 existing 2D datasets with three open-source 3D datasets for joint training using a total of 17 datasets.
The datasets we used are as follows:

\begin{itemize}
\item 3 whole-body datasets: COCO-Wholebody~\cite{zoomnet, xu2022zoomnas}, UBody~\cite{lin2023one}, Halpe~\cite{alphapose}
\item 6 human body datasets: AIC~\cite{fulgeri2019can}, CrowdPose~\cite{li2018crowdpose}, MPII~\cite{andriluka14cvpr}, sub-JHMDB~\cite{Jhuang:ICCV:2013}, PoseTrack18~\cite{xiao2018simple}, Human-Art~\cite{ju2023human}
\item 4 face datasets: WFLW~\cite{wayne2018lab}, 300W, COFW, LaPa~\cite{liu2020new}
\item 1 hand dataset: InterHand~\cite{Moon_2020_ECCV_InterHand2.6M}
\item 3 3D whole-body point datasets: H3WB~\cite{Zhu_2023_ICCV}, UBody~\cite{lin2023one}, DNA-rendering~\cite{2023dnarendering}
\end{itemize}

For detailed mapping relationships, please refer to the appendix.
We carried out a two-stage distillation of RTMW according to the process proposed by DWPose~\cite{yang2023effective}, and the relevant hyperparameters followed the settings in DWPose~\cite{yang2023effective}.

\subsection{Results}

\paragraph{COCO-Wholebody}
We validate the proposed RTMW model on the whole-body pose estimation task with COCO-WholeBody~\cite{xu2022zoomnas, jin2020whole} V1.0 dataset. As shown in Table~\ref{tab:compare_cocowhole}, RTMW achieves superior performance and well balances accuracy and complexity. In addition, our proposed RTMW3D also demonstrates good performance on COCO-WholeBody.

\begin{table*}[h]
    \centering
    \caption{Whole-body pose estimation results on COCO-WholeBody~\cite{xu2022zoomnas, jin2020whole} V1.0 dataset. ``*'' denotes the model is pre-trained on AIC+COCO. ``$^\dag$'' denotes the model in trained on the combined dataset introduced at \ref{sec:Experiments}.}
    \vspace{5pt}
    \resizebox{0.8\linewidth}{!}{
    \begin{tabular}{c|l|c|cc|cc|cc|cc|cc}
         \toprule
         & Method & Input Size & \multicolumn{2}{c|}{whole-body} & \multicolumn{2}{c|}{body} & \multicolumn{2}{c|}{foot} & \multicolumn{2}{c|}{face} & \multicolumn{2}{c}{hand} \\
         \cmidrule{4-13}
         &        &            & AP & AR  & AP & AR & AP & AR & AP & AR & AP & AR   \\
         \midrule
         & DeepPose~\cite{Toshev_2014_CVPR} & 384$\times$288  & 33.5 & 45.6 & 42.7 & 58.3 & 9.9 & 36.9 & 64.9 & 69.7 & 40.8 & 58.0 \\
         & SimpleBaseline~\cite{xiao2018simple} & 384$\times$288 & 57.3 & 67.1 & 66.6 & 74.7 & 63.5 & 76.3 & 73.2 & 81.2 & 53.7 & 64.7 \\
         & HRNet~\cite{SunXLW19}  & 384$\times$288 & 58.6 & 67.4 & 70.1 & 77.3 & 58.6 & 69.2 & 72.7 & 78.3 & 51.6 & 60.4 \\
       Top-Down  & PVT~\cite{WenhaiWang2021PyramidVT} & 384$\times$288 & 58.9 & 68.9 & 67.3 & 76.1 & 66.0 & 79.4 & 74.5 & 82.2 & 54.5 & 65.4 \\
       Methods  & FastPose50-dcn-si~\cite{alphapose} &  256$\times$192 & 59.2 & 66.5 & 70.6 & 75.6 & 70.2 & 77.5 & 77.5 & 82.5 & 45.7 & 53.9 \\
         & ZoomNet~\cite{zoomnet} & 384$\times$288 & 63.0 & 74.2 & 74.5 & 81.0 & 60.9 & 70.8 & 88.0 & 92.4 & 57.9 & 73.4 \\
         & ZoomNAS~\cite{xu2022zoomnas} & 384$\times$288 & 65.4 & 74.4 & 74.0 & 80.7 & 61.7 & 71.8 & 88.9 & 93.0 & 62.5 & 74.0 \\
         & DWPose-l~\cite{yang2023effective} & 384$\times$288 & 66.5 & 74.3 & 72.2 & 78.9 & 70.4 & 81.6 & 88.7 & 92.1 & 62.1 & 72.0 \\
         \midrule
         & RTMPose-m* & 256$\times$192 & 58.2 & 67.4 & 67.3 & 75.0 & 61.5 & 75.2 & 81.3 & 87.1 & 47.5 & 58.9 \\
         & RTMPose-l* & 256$\times$192 & 61.1 & 70.0 & 69.5 & 76.9 & 65.8 & 78.5 & 83.3 & 88.7 & 51.9 & 62.8 \\
       RTMPose~\cite{https://doi.org/10.48550/arxiv.2303.07399}  & RTMPose-l* & 384$\times$288 & 64.8 & 73.0 & 71.2 & 78.1 & 69.3 & 81.1 & 88.2 & 91.9 & 57.9 & 67.7 \\
         & RTMPose-x* & 384$\times$288 & 65.2 & 73.2 & 71.2 & 78.0 & 68.1 & 80.4 & 89.0 & 92.2 & 59.3 & 68.7 \\
         & RTMPose-x* & 384$\times$288 & 65.3 & 73.3 & 71.4 & 78.4 & 69.2 & 81.0 & 88.9 & 92.3 & 59.0 & 68.5 \\
         \midrule
      RTMW3D   & RTMW3D-l$^\dag$ & 384$\times$288 & 67.8 & 75.5 & 74.4 & 80.7 & 77.1 & 86.7 & 88.2 & 91.9 & 61.7 & 70.6 \\
         & RTMW3D-x$^\dag$ & 384$\times$288 & 68.0 & 75.9 & 73.9 & 80.5 & 76.6 & 86.8 & 88.6 & 92.1 & 62.7 & 72.2  \\
         \midrule
         & RTMW-m$^\dag$ & 256$\times$192 & 58.0 & 67.3 & 67.6 & 74.7 & 67.1 & 79.4 & 78.3 & 85.4 & 49.1 & 60.4 \\
         & RTMW-l$^\dag$ & 256$\times$192 & 66.2 & 74.6 & 74.3 & 80.7 & 76.3 & 86.8 & 83.4 & 88.9 & 59.8 & 70.1 \\
      RTMW   & RTMW-x$^\dag$ & 256$\times$192 & 67.2 & 75.2 & 74.6 & 80.8 & 77.0 & 86.9 & 84.4 & 89.6 & 61.0 & 71.0 \\
         & RTMW-l$^\dag$ & 384$\times$288 & 70.1 & 78.0 & 76.1 & 82.4 & 79.3 & 88.5 & 88.4 & 92.1 & 66.3 & 75.2 \\
         & RTMW-x$^\dag$ & 384$\times$288 & 70.2 & 78.1 & 76.3 & 82.6 & 79.6 & 88.8 & 88.4 & 92.3 & 66.4 & 75.5 \\
         \bottomrule
    \end{tabular}}
    \label{tab:compare_cocowhole}
\end{table*}

\paragraph{H3WB}
H3WB~\cite{Zhu_2023_ICCV} is currently the only open-source 3D whole-body pose estimation dataset that provides a test set. We evaluated the performance of RTMW3D on the test set of H3WB, and the results are shown in Table~\ref{tab:compare_h3wb}. Similarly, RTMW3D has achieved excellent performance on the H3WB dataset.

\begin{table}[h]
\begin{center}
 \caption{Whole-body 3D pose estimation results on H3WB~\cite{Zhu_2023_ICCV} dataset. ``*'' denotes the results are borrowed from H3WB paper~\cite{Zhu_2023_ICCV}.}\label{tab:compare_h3wb}
    \vspace{3pt}
    \resizebox{0.9\linewidth}{!}{
		\begin{tabular}{l|c|c}
			\toprule
                Method & Input Size & MPJPE \\
	
			\midrule
                CanonPose~\cite{Wandt2021Canonpose} with 3D supervision* & N/A & 0.117 \\
                Large SimpleBaseline*~\cite{xiao2018simple} & N/A & 0.112 \\
                JointFormer*~\cite{lutz2021joint} & N/A & 0.088 \\
                RTMW3D-l & 384x288 & 0.056 \\
                RTMW3D-x & 384x288 & 0.057 \\
			\bottomrule
		\end{tabular}}
	\end{center}
\end{table}

\paragraph{Inference speed}
As our primary focus is on the development of a real-time model, the inference speed emerges as a critical performance metric. The evaluation of RTMW models' inference speed is detailed in the accompanying table~\ref{tab:onnx_speed}. Although RTMW, which incorporates an additional module relative to RTMPose, exhibits a marginally reduced inference speed, it offers a substantial enhancement in accuracy. This trade-off is deemed justifiable in the context of our objectives.

\begin{table}[h]
\caption{Inference speed on CPU. RTMPose models are deployed and tested using ONNXRuntime. Flip test is not used in this table. }\label{tab:onnx_speed}
	\begin{center}
    \resizebox{\linewidth}{!}{
		\begin{tabular}{c|l|c|c|c|c}
		    \toprule
			\multicolumn{2}{c|}{Results} & Input Size & GFLOPs & AP & CPU(ms)   \\
			\midrule
			\multirow{5}{*}
             & HRNet-w32+DARK & $256 \times 192$ & 7.72& 57.8 & 39.051 \\ 
             & RTMPose-m & $256 \times 192$ & 2.22 & 59.1 & \textbf{13.496} \\ 
             & RTMPose-l & $256 \times 192$ & 4.52 & 62.2 & 23.410 \\ 
       COCO- & HRNet-w48+DARK & $384 \times 288$ & 35.52 & 65.3 & 150.765 \\ 
WholeBody~\cite{zoomnet} & RTMPose-l & $384 \times 288$ & 10.07 & 66.1 & 44.581  \\ 
            \cmidrule{2-6}
            & RTMW-l    & $256 \times 192$ & 7.9 & 66.0 & 35.322  \\
            & RTMW-l    & $384 \times 288$ & 17.7 & \textbf{70.1} & 47.618  \\
			\bottomrule
		\end{tabular}}
	\end{center}
\end{table}

\begin{table*}[h]
    \centering
    \caption{Ablation on RTMW. Models are trained on COCO-WholeBody+UBody+Halpe.}
    \vspace{5pt}
    \scalebox{0.9}{
    \begin{tabular}{l|c|c|c|c|c|c|c|l}
         \toprule
             & PAFPN & HEM & GAU & Whole & Body & Foot & Face & Hand  \\
         \midrule
         RTMPose-l &  &  & \checkmark & 63.8 & 71.7 & 71.5 & 83.3 & 56.4 \\
         \midrule
         w/ PAFPN  & \checkmark & & \checkmark & 64.3 (\textcolor{blue}{+0.5}) & 71.8 (\textcolor{blue}{+0.1}) & 71.3 (\textcolor{red}{-0.2}) & 84.0 (\textcolor{green}{+0.7}) & 57.7 (\textcolor{green}{+1.3}) \\
         w/ HEM    & & \checkmark & \checkmark & 64.1 (\textcolor{blue}{+0.3}) & 72.0 (\textcolor{blue}{+0.3}) & 71.9 (\textcolor{blue}{+0.4}) & 83.4 (\textcolor{blue}{+0.1}) & 57.1 (\textcolor{green}{+0.7}) \\
         w/ PAFPN + HEM & \checkmark & \checkmark & \checkmark & 65.3 (\textcolor{green}{+1.5}) & 72.7 (\textcolor{green}{+1.0}) & 74.8 (\textcolor{green}{+3.3}) & 84.0 (\textcolor{green}{+0.7}) & 59.9 (\textcolor{green}{+3.5}) \\
        w/o GAU  & \checkmark & \checkmark &  & 64.8 (\textcolor{green}{+1.0}) & 71.9 (\textcolor{blue}{+0.2}) & 72.2 (\textcolor{green}{+0.7}) & 84.1 (\textcolor{green}{+0.8}) & 59.5 (\textcolor{green}{+3.1}) \\
        \bottomrule
    \end{tabular}
    }
    \label{tab:rtmw_ablation}
\end{table*}

\subsection{Ablation study}

\subsubsection{Ablation on RTMW}

We tested the impact of each module of RTMW on the performance, and the results are shown in table~\ref{tab:rtmw_ablation}. The models are trained on COCO-Wholebody~\cite{xu2022zoomnas, jin2020whole}, Ubody~\cite{lin2023one}, and Halpe~\cite{alphapose} datasets, and the input shape is $256\times192$. The experimental results indicate that the PAFPN and HEM modules we employed can significantly enhance the prediction accuracy, particularly for body parts like hands and feet with lower input resolutions, where the improvement in predictive accuracy is substantial.

\subsection{Visualization Results}
\begin{figure*}[h]
    \centering
    \begin{minipage}{0.3\linewidth}
        \vspace{3pt}
        \centerline{\includegraphics[width=0.9\textwidth]{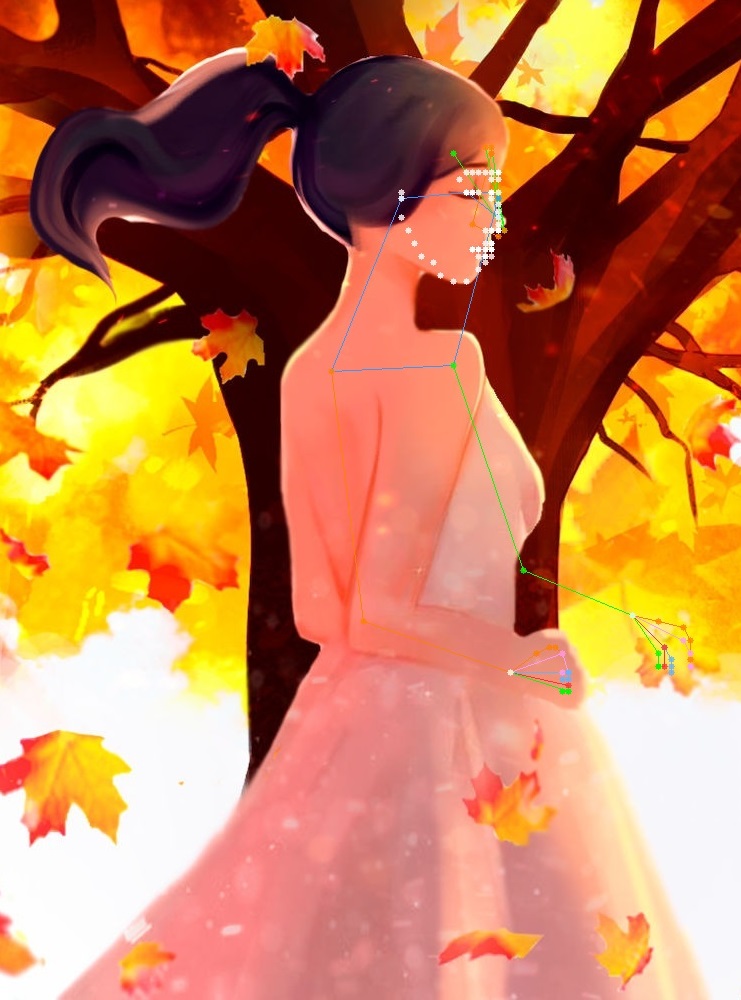}}
        \centerline{\fontsize{7pt}{5pt}\selectfont HRNet~\cite{SunXLW19}}
    \end{minipage}
    \begin{minipage}{0.3\linewidth}
        \vspace{3pt}
        \centerline{\includegraphics[width=0.9\textwidth]{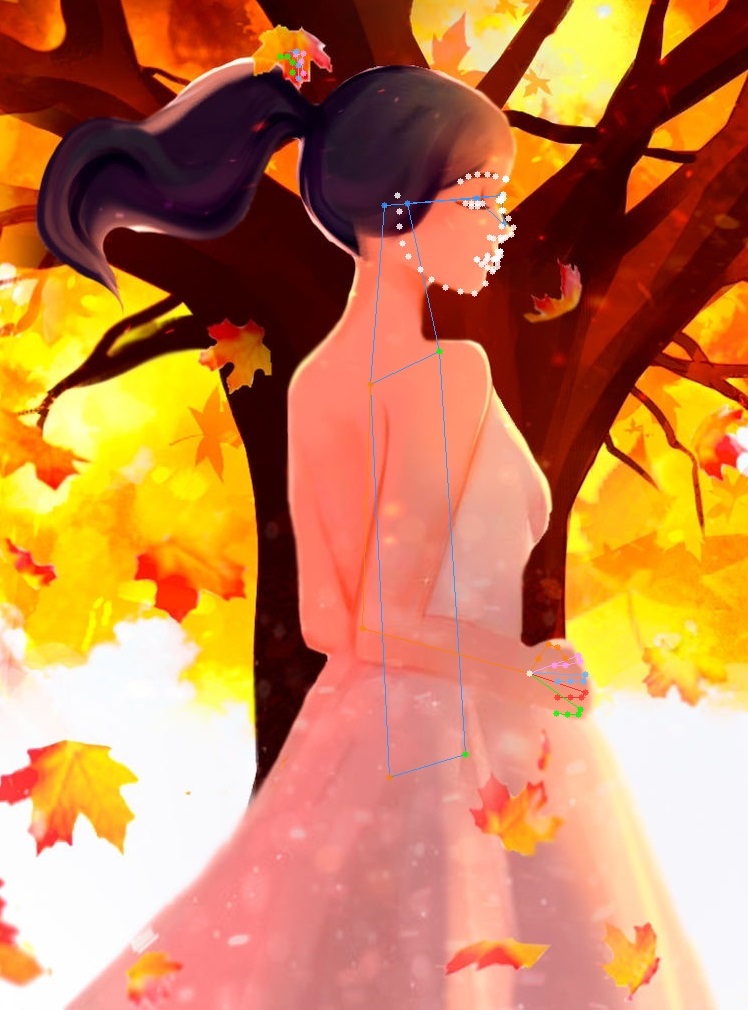}}
        \centerline{\fontsize{7pt}{5pt}\selectfont DWPose~\cite{yang2023effective}}
    \end{minipage}
    \begin{minipage}{0.3\linewidth}
        \vspace{3pt}
        \centerline{\includegraphics[width=0.87\textwidth]{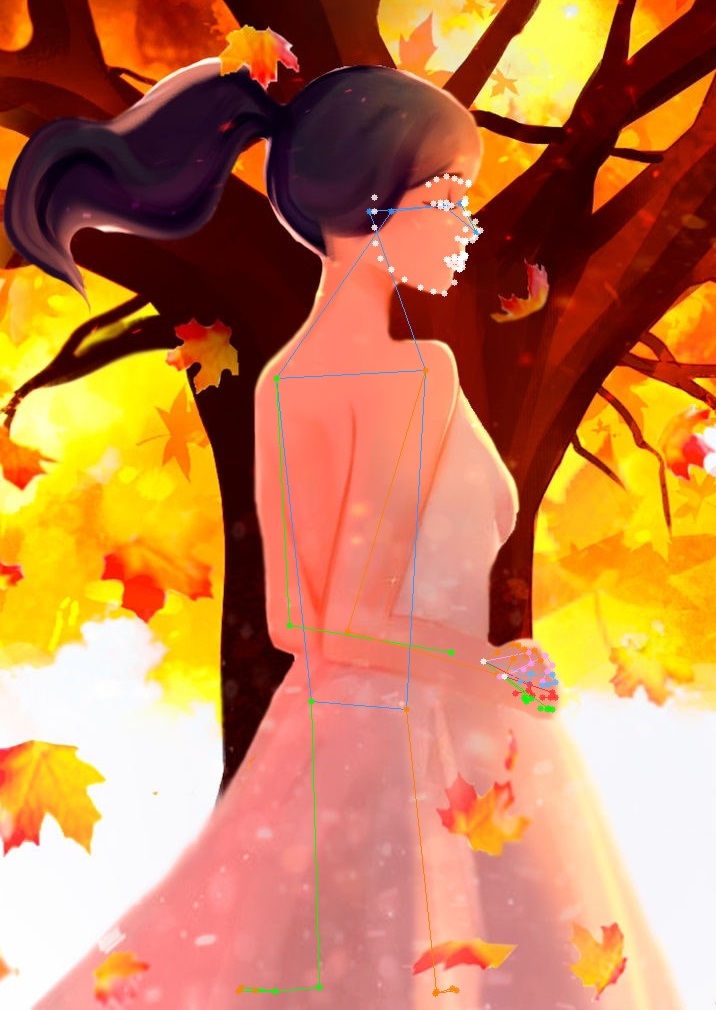}}
        \centerline{\fontsize{7pt}{5pt}\selectfont ours}
    \end{minipage}
    \vspace{5pt}
    \caption{2D visualization results}
    \label{fig:2d_visualization}
\end{figure*}

\begin{figure}
    \centering
    \includegraphics[width=0.8\linewidth]{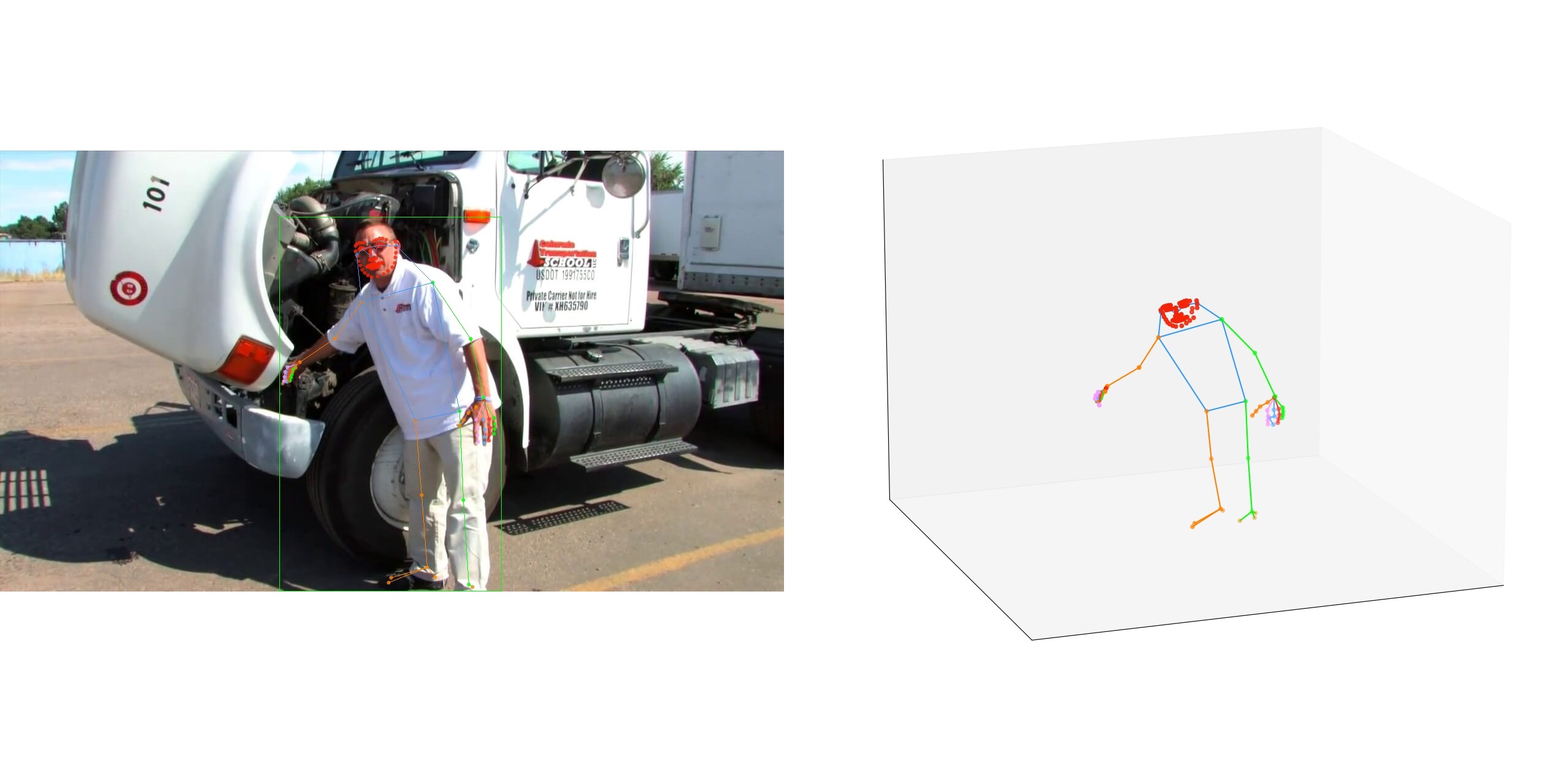}
    \includegraphics[width=0.8\linewidth]{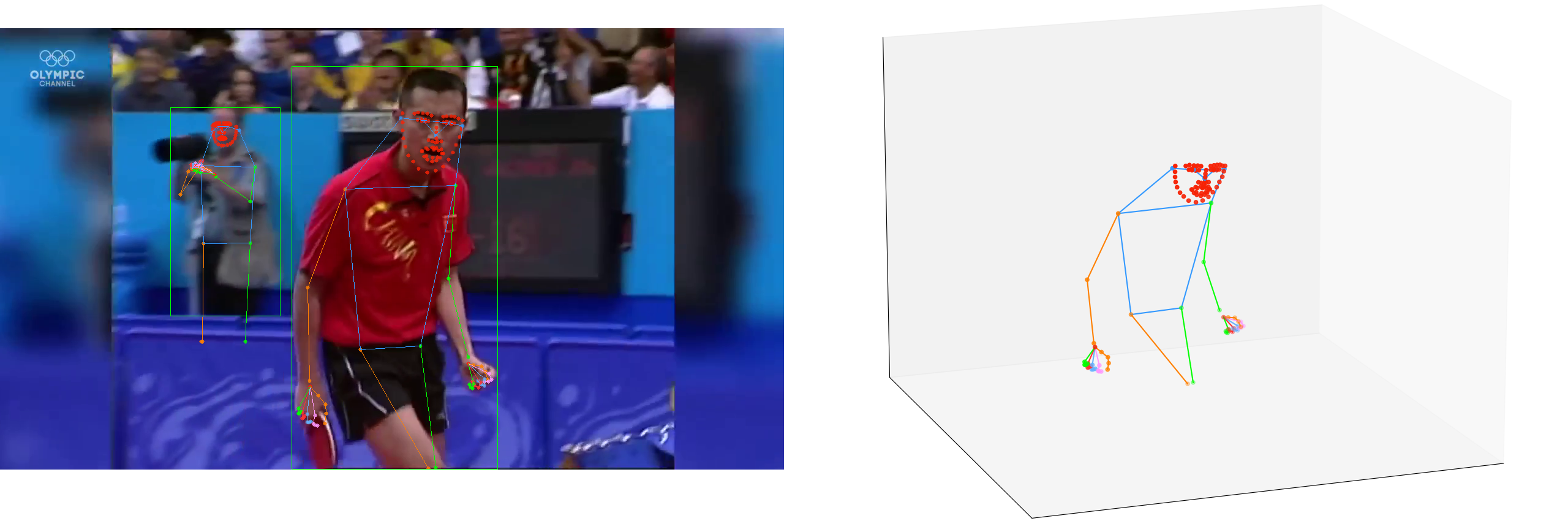}
    \includegraphics[width=0.8\linewidth]{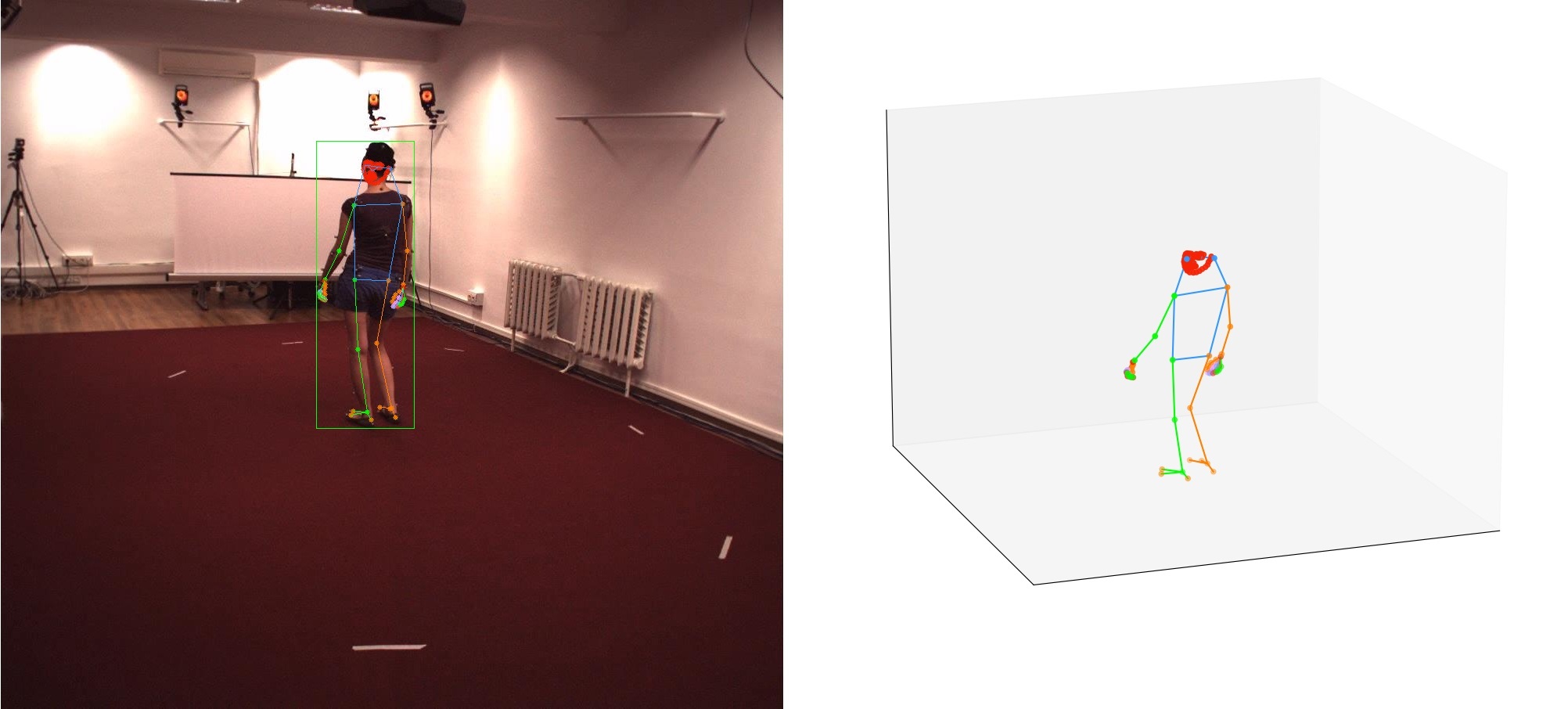}
    \vspace{5pt}
    \caption{RTMW3D inference results \fontsize{8pt}{1pt}\selectfont{(The 2D and 3D keypoints are both obtained from the RTMW3D model after a single inference.})}
    \label{fig:3d_visualization}
\end{figure}

Figure~\ref{fig:2d_visualization} and Figure~\ref{fig:3d_visualization} present visual representations of the inference outcomes for the RTMW and RTMW3D models. These illustrations demonstrate that the performance of both RTMW and RTMW3D models is commendably high, corroborating the quantitative assessment data.

\section{Conclusion}\label{sec:Conclusion}


This paper expands the existing scholarly work by critically examining the intricacies and challenges inherent in whole-body pose estimation. Leveraging our established RTMPose model as a foundation, we introduce an enhanced, high-performance model, RTMW/RTMW3D, for real-time whole-body pose estimation. Our model has demonstrated unparalleled performance among all open-source alternatives and possesses a distinct monocular 3D pose estimation capability. We anticipate that the proposed algorithm and its open-source availability will address several practical requirements within the industry for robust pose estimation solutions.

{\small
\bibliographystyle{ieee_fullname}
\bibliography{egbib}
}

\end{document}